\documentclass[runningheads]{llncs}
\usepackage[T1]{fontenc}
\usepackage{graphicx}
\usepackage{array}
\newcolumntype{P}[1]{>{\centering\arraybackslash}p{#1}}
\usepackage{amsmath}
\usepackage{multirow}
\begin{document}
\title{Multimodal Misinformation Detection in a South African Social Media Environment}
\titlerunning{South African Misinformation Detection}
%
\author{Amica De Jager\inst{1}\orcidID{0000-0002-4173-3417} \and
Vukosi Marivate\inst{2}\orcidID{0000-0002-6731-6267} \and
Abiodun Modupe\inst{3}\orcidID{0000-0002-9732-6466}}
\authorrunning{A. De Jager et al.}
%
\institute{University of Pretoria, Pretoria, 0028, South Africa \\
\email{amicadejager@gmail.com} \and
University of Pretoria, Pretoria, 0028, South Africa \\
\email{vukosi.marivate@cs.up.ac.za}\and
University of Pretoria, Pretoria, 0028, South Africa \\
\email{abiodun.modupe@cs.up.ac.za}}
\maketitle              
\begin{abstract}
The world is witnessing a growing epidemic of misinformation. Misinformation can have severe impacts on society across multiple domains: including health, politics, security, the environment, the economy and education. With the constant spread of misinformation on social media networks, a need has arisen to continuously assess the veracity of digital content. This need has inspired numerous research efforts on the development of misinformation detection (MD) models. However, many models do not use all information available to them and existing research contains a lack of relevant datasets to train the models, specifically within the South African social media environment. The aim of this paper is to investigate the transferability of knowledge of a MD model between different contextual environments. This research contributes a multimodal MD model capable of functioning in the South African social media environment, as well as introduces a South African misinformation dataset. The model makes use of multiple sources of information for misinformation detection, namely: textual and visual elements. It uses bidirectional encoder representations from transformers (BERT) as the textual encoder and a residual network (ResNet) as the visual encoder. The model is trained and evaluated on the Fakeddit dataset and a South African misinformation dataset. Results show that using South African samples in the training of the model increases model performance, in a South African contextual environment, and that a multimodal model retains significantly more knowledge than both the textual and visual unimodal models. Our study suggests that the performance of a misinformation detection model is influenced by the cultural nuances of its operating environment and multimodal models assist in the transferability of knowledge between different contextual environments. Therefore, local data should be incorporated into the training process of a misinformation detection model in order to optimize model performance.
\keywords{Misinformation \and Disinformation \and Fake News Detection \and Multimodal \and Natural Language Processing \and Transfer Learning \and Deep Learning \and Social Media \and South African.}
\end{abstract}
\section{Introduction}

Misinformation can have severe impacts on society across multiple domains, including health, politics, security, the environment, the economy and education. These impacts are significantly amplified on social media networks due the effortless content creation capabilities, easy accessibility to global audiences, and the speed and scope at which information can be diffused across networks in real-time. Due to the aforementioned characteristics of social media networks, misinformation is granted an enormous potential to cause real harm, within minutes, for millions of users~\cite{ref_figueira}. For example, the sharing of misinformation on social media networks was used in South Africa to encourage civil unrest, resulting in city-wide looting and riots, known as the Durban Riots during 2021~\cite{ref_lapping}. Another example includes how misinformation was used by domestic and foreign actors to influence public opinion, undermining democracy, in the 2016 American presidential elections~\cite{ref_bentzen}. Misinformation was also used extensively during the COVID-19 pandemic to cause confusion, encourage risk-taking behaviours and public mistrust in health authorities, thereby undermining the public health response~\cite{ref_who}. Overall, misinformation is a global issue that warrants the need to monitor, assess and regulate the veracity of digital content. Thus, inspiring research in the pursuit of mitigating this crisis through the development of misinformation detection models.

In the subsections to follow, misinformation and the subtypes of misinformation will be defined. Additionally, different groups of misinformation detection models will be outlined, followed by a discussion on concept drift and its applicability to misinformation detection models for the South African context.

\subsection{Misinformation} \label{sec_types}
Misinformation refers to false or inaccurate information. It should be noted that there are several terms relating to misinformation which may be confusing, such as: fake news and disinformation. Fake news refers to misinformation that is distributed in the form of news articles. Both misinformation and disinformation refer to false or inaccurate information, however the distinction between these terms is whether the information was created with the intent to deceive. Misinformation often refers to the unintentional instances, while disinformation, the intentional instances where the information has ill intent and is often a deliberate attack to a particular group~\cite{ref_wu}. However, since it is difficult for researchers to determine intent, misinformation is used throughout this article as an umbrella term to include all false or inaccurate information.

Misinformation can be classified into different categories according to the type of content being created or shared, the motivations of the content creator and the methods used to share the content. Wardle~\cite{ref_wardle} defined seven categories of misinformation, based on a scale measuring the content creator’s intent to deceive, namely: satire or parody, false connection, misleading content, false context, imposter content, manipulated content and fabricated content. These categories are defined in Table~\ref{tab_wardle}. These same categories were later used by Nakamura, Levy and Wang~\cite{ref_nakamura} in creating a misinformation dataset, Fakeddit, to support in the development of misinformation detection models. Misinformation detection is often treated as a binary classification task (i.e., true or false), grouping all types of misinformation into a single parent class. However, misinformation detection can be extended into a multi-class classification problem, thus accounting for the different types of misinformation, as described in Table~\ref{tab_wardle}.

\begin{table}
\centering
\caption{Wardle's seven categories of misinformation~\cite{ref_wardle}.}\label{tab_wardle}
\begin{tabular}{|P{2.5cm}|P{3.2cm}|P{6cm}|}
\hline
{\bfseries Measure of Intent to Deceive} &  {\bfseries Misinformation Category} & {\bfseries Definition}\\
\hline
1 &  Satire or Parody & The meaning of the content is twisted or misinterpreted in a satirical or humorous way.\\ \hline
2 & False Connection & Headlines, visuals or captions do not support the content.\\ \hline
3 & Misleading Content & Misleading use of information to frame an issue or individual.\\ \hline
4 & False Context & Genuine content is shared with false contextual information.\\ \hline
5 & Imposter Content & Genuine sources are impersonated.\\ \hline
6 & Manipulated Content & Genuine information or imagery is manipulated to deceive (e.g., photo editing).\\ \hline
7 & Fabricated Content & New content is created that is completely false, designed to deceive and cause harm.\\
\hline
\end{tabular}
\end{table}

\subsection{Misinformation Detection Models}

Before the development of misinformation detection (MD) models, misinformation detection was performed solely by professional fact-checkers. Professional fact-checking as a process involves manually reading through content, researching and comparing knowledge extracted from the content with factual information in that particular domain, and making a decision on the content’s authenticity based on the subsequent findings. Although manual fact-checking plays a critical role in misinformation detection, it is not sufficient when analysing the immense volume and variety of digital content that is being rapidly diffused on social media networks~\cite{ref_capuano,ref_zhang}. For example, text created on high frequency social media platforms can be upwards of one million sentences every few minutes, making it impossible for professional fact-checkers to read through and validate all the information. Therefore, MD models, which involves leveraging the benefits of machine learning, are developed to automate the process, assisting fact-checkers in the task of misinformation detection.

\begin{figure}
\includegraphics[width=\textwidth]{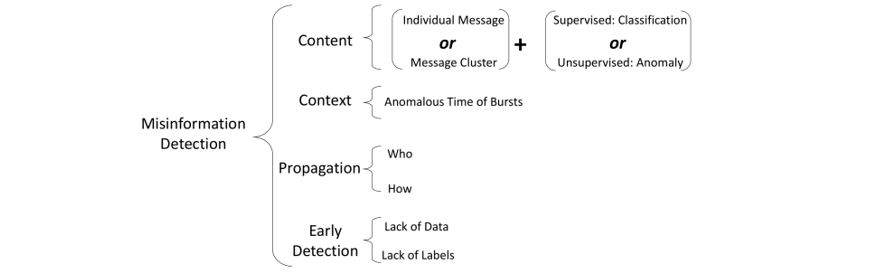}
\caption{An overview of the categorization of misinformation detection models~\cite{ref_wu}.} \label{fig_wu}
\end{figure}

Misinformation detection models can be categorized based on the type of information the model uses for detection. These categories, displayed in Figure~\ref{fig_wu}, include: content-based, context-based, propagation-based and early detection. Content-based MD models detect misinformation directly on the content of a sample, which forms part of a training dataset. This involves using text, audio, image, video, or any combination thereof to generate the training data used in a model. Context-based MD models, on the other hand, detect misinformation based on the contextual information surrounding a sample on a social media network. These types of models work by using information such as a sample’s geolocation, the time that a sample is created, the account, profile or source responsible for a sample and the reaction of network users to a sample. Propagation-based MD models detect misinformation based on the propagation patterns of a sample over a network, also referred to as information diffusion. Propagation-based MD models also consider the users who share a sample and the speed at which a sample is diffused. Early detection models focus on detecting misinformation at an early stage before the content becomes viral. Models in this category need to be robust in order to handle the two major challenges that arise in the task of early misinformation detection, namely: a lack of data and a lack of labels~\cite{ref_wu}. Early detection models are either one of the aforementioned models (content-based, context-based or propagation-based), or a hybrid between these models. This paper focuses on content-based methods, as these models consider the veracity of the actual content, as opposed to the circumstances surrounding the content on social media networks.

Misinformation detection models can also be categorized according to the quantity of information sources that are utilized. If a single source of information is used for detection (i.e., text), then the model is referred to as a unimodal model. In contrast, if multiple sources of information are utilized (i.e., text and image), then the model is referred to as a multimodal model. Since the majority of content shared on social media networks include both text and images, unimodal models tend to be insufficient for misinformation detection on these platforms. Thus, multimodal models, which are capable of processing both textual and visual inputs, are the preferred choice for improving the classification of misinformation on social media networks~\cite{ref_palani}.

\subsection{Concept Drift} \label{sec_concept}

Misinformation detection models are vulnerable to concept drift. Concept drift occurs when the underlying relationship between the input data and the target variable changes over time~\cite{ref_brownlee}. In context of misinformation detection, this can occur when many labelled samples from verified misinformation content becomes outdated, with the introduction of newly developed content. For example, a MD model that is trained on misinformation data before the COVID-19 pandemic may struggle to classify misinformation during the pandemic~\cite{ref_raza}. Another example is a MD model that is trained on a misinformation dataset, originating from a specific country or cultural context, may struggle to classify misinformation when applied to a different country or cultural context because the nuance of that setting might not be applicable for another region or culture. Therefore, it is important to consider the contextual environment in which the MD model is deployed and to ensure the model is trained on relevant, updated datasets.

\subsection{Misinformation Detection in South Africa}

\begin{figure}
\includegraphics[width=\textwidth]{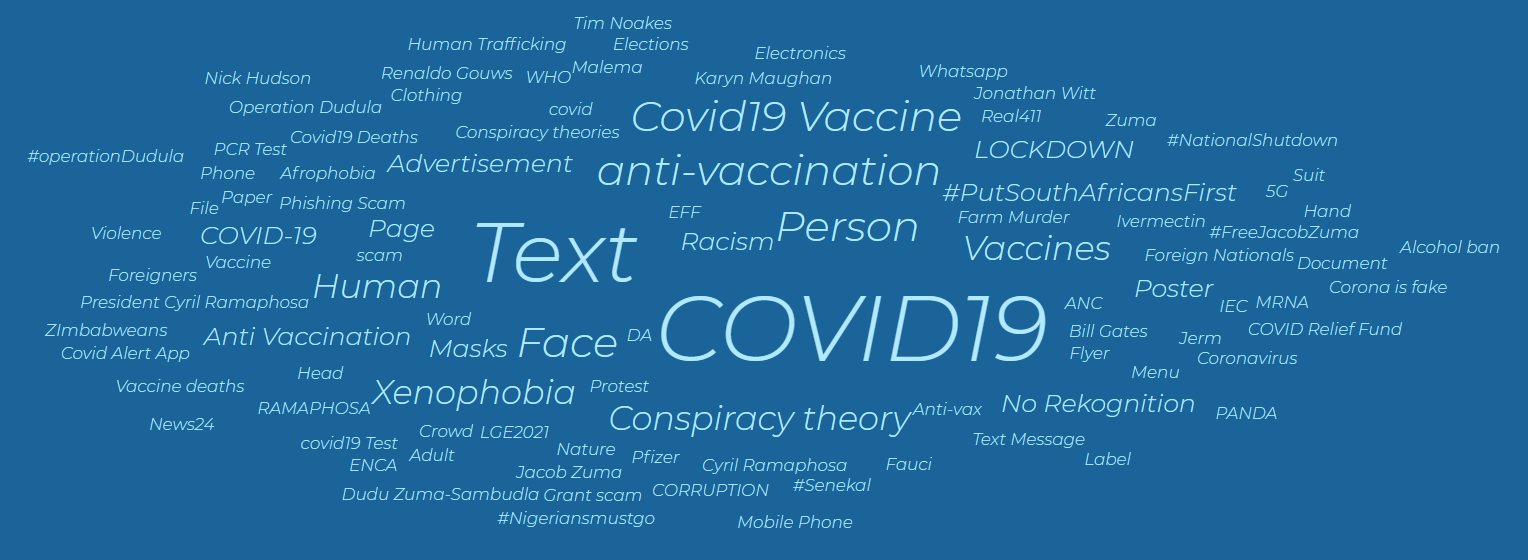}
\caption{Word cloud of most common Real411 complaints by tag~\cite{ref_real411}.} \label{fig_real411}
\end{figure}

Due to the potential influence of concept drift on the performance of misinformation detection models, it is important to discuss misinformation in the South African context. There are, currently, no South African misinformation datasets available to support in the development of MD models, specifically for use in South Africa. However, there is a manual fact-checking program, called Real411~\cite{ref_real411}. Real411 hosts a website on which members of the South African community can submit complaints flagging potential cases of misinformation, hate speech, incitement to violence or harassment. Complaints typically include both textual and visual elements. The most common complaints are displayed, by tag, in Figure~\ref{fig_real411}. Once a complaint is submitted, the Real411 Digital Complaints Committee (DCC) assesses the veracity of the flagged content and if the complaint is proven valid, the DCC initiates appropriate action, which may include: seeking assistance from the relevant online platforms, being referred to the South African Human Rights Commission (SAHRC) or reported to the South African Police Service (SAPS). For the purposes of this research, Real411 has made their data available, to become South Africa's first misinformation dataset, in order to assist in developing MD models capable of functioning in the South African social media environment.

\section{Related Works}

Yuan, et al.,~\cite{ref_yuan} developed an Event Adversarial Neural Network (EANN) for multimodal misinformation detection. The model consists of a modified convolutional neural network (CNN) for textual feature extraction and a pre-trained, 19-layer CNN from Visual Geometry Group (VGG19) for visual feature extraction. An event discriminator is also used in the framework to remove all event-specific features, maintaining the shared features among events. The EANN model is trained and evaluated on a Twitter and Weibo misinformation dataset. Khattar, Goud, Varma and Gupta~\cite{ref_khattar} developed a Multimodal Variational Autoencoder (MVAE) which makes use of a bidirectional long-short term memory (Bi-LSTM) encoder for textual feature extraction and VGG19 for visual feature extraction. The MVAE model learns a joint latent vector by optimizing the process of encoding a text-image sample and decoding the result to reconstruct the original sample. This learnt representation is then used for binary classification. Like EANN, MVAE is also trained and evaluated on a Twitter and Weibo misinformation dataset. Singhal, et al.,~\cite{ref_singhal} developed a multimodal model, SpotFake+, which was one of the first misinformation detection models to leverage the benefits of transfer learning by making use of the pre-trained large language model (LLM), XLNet, for textual feature extraction and VGG19 for visual feature extraction. The SpotFake+ model is trained and evaluated on the Politifact and Gossipcop datasets from the FakeNewsNet repository~\cite{ref_shu}. Politifact incorporates misinformation from the political domain, while Gossipcop incorporates misinformation from the entertainment domain. Alonso-Bartolome and Segura-Bedmar~\cite{ref_alonso} developed a multimodal model which makes use of CNN architecture for both textual and visual feature extraction. The model is trained and evaluated using the Fakeddit dataset which is a cross-domain, multi-class misinformation dataset~\cite{ref_nakamura}. The Fakeddit dataset is one of the largest misinformation datasets available with over one million samples. The model achieves state-of-the-art results with an overall accuracy of 87\% at detecting misinformation. Palani, Elango and Viswanathan~\cite{ref_palani} developed a multimodal model, CB-Fake, which makes use of bidirectional encoder representations from transformers (BERT), a pre-trained LLM, for textual feature extraction and a capsule neural network (CapsNet) for visual feature extraction. The CB-Fake model is also trained and evaluated on the Politifact and Gossipcop datasets, achieving current state-of-the-art results in misinformation detection. Zhou, Ying, Qian, Li, and Zhang~\cite{ref_zhou} developed a multimodal misinformation detection network (FND-CLIP) based on contrastive language-image pretraining (CLIP). The FND-CLIP framework uses pre-trained BERT for textual feature extraction, pre-trained residual neural network (ResNet) for visual feature extraction and CLIP encoders to allow for finer-grained feature fusion between the textual and visual features. The framework also incorporates a modality-wise attention module to adaptively reweight and aggregate the features. FND-CLIP is trained and evaluated on Weibo, Politifact and Gossipcop datasets, outperforming state-of-the-art multimodal misinformation detection models. A review of literature highlights a lack of research of misinformation detection within a South African context and all existing datasets originate from America, Europe or Asia. Therefore, the purpose of this paper is to contribute a multimodal MD model capable of detecting misinformation in the South African social media environment, as well as introduces the first South African misinformation dataset.

\section{Methodology}

This section formulates the problem and discusses the proposed framework of a multimodal misinformation detection model used in a South African social media environment. The section concludes with a description of the experimental setup.

\subsection{Problem Formulation} \label{sec_prob}



This article presents two tasks. The aim of Task 1 is to evaluate the proposed model on a well-known dataset in three different classification settings: binary, 3-class and 6-class classification. The purpose of classification is to identify whether a sample, sourced from a social media network, is misinformation or not. Task 2 extends Task 1 by comparing the model’s performance on a local South African dataset, when trained on non-local samples, a mix of non-local and local samples and local samples. The second task investigates the effect of knowledge transferability of the MD model between different contextual environments. It is also important to note that during both tasks, the proposed multimodal MD model is compared with both textual and visual unimodal models. This is performed in order to investigate the contribution of multimodality to the task of misinformation detection. For the purpose of this article, misinformation detection is modelled as a binary classification problem, as presented in Equation~\ref{equ_1}.

\begin{equation}
F(S) = 
\begin{cases}
  0 & \mbox{ if $S$ is not misinformation}\\
  1 & \mbox{ if $S$ is misinformation}
\end{cases}
\label{equ_1}
\end{equation}

Let $S = \{s_1, s_2, \ldots, s_n\}$ represent a misinformation dataset of $n$ samples, with $s_i = t_i \cup v_i$, where $t_i$ and $v_i$ are the text and image extracts associated with the $i$th sample, respectively. Let $Y = \{y_1, y_2, \ldots, y_n\}$ represent the set of ground-truth labels for the misinformation dataset, such that the $i$th label, for the $i$th sample, is given by: $y_i \in \{0,1\}$. Then, a multimodal MD model $F : S \rightarrow Y$, is represented in Equation~\ref{equ_1}.

\subsection{Proposed MMiC Model}

\begin{figure}
\includegraphics[width=\textwidth]{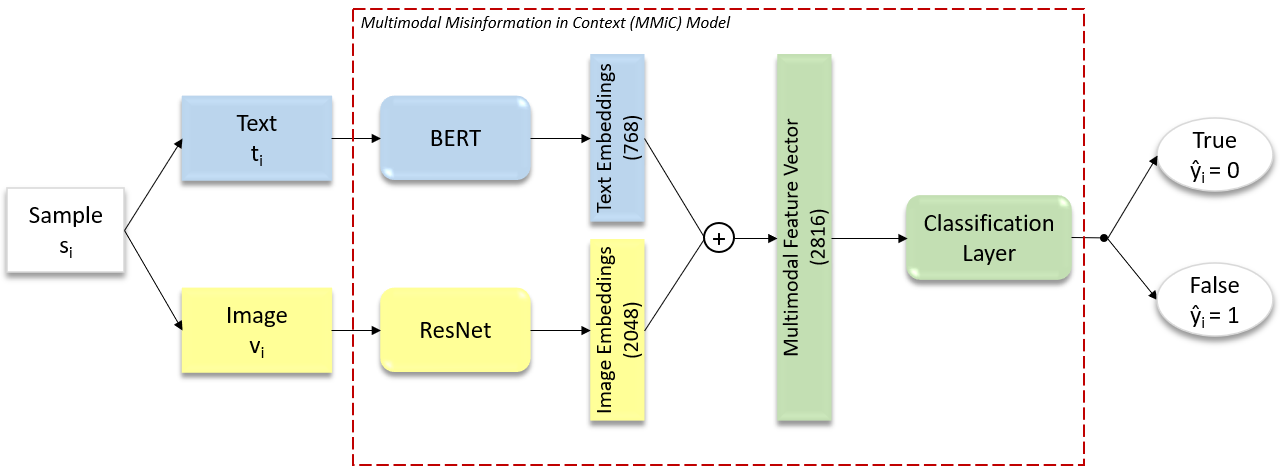}
\caption{Diagram of proposed MMiC model for misinformation detection.} \label{fig_model}
\end{figure}

This article proposes a multimodal misinformation detection model that is capable of detecting misinformation in the South African social media environment. The Multimodal Misinformation-in-Context (MMiC) model uses pre-trained bidirectional encoder representations from transformers (BERT) model for textual feature extraction and pre-trained residual neural network (ResNet) model for visual feature extraction, as these encoders are used by the current state-of-the-art FND-CLIP model~\cite{ref_zhou}. The MMiC model is outlined in Figure~\ref{fig_model}.

A misinformation dataset is pre-processed before entering the MMiC model framework. After pre-processing is complete, the multimodal sample enters the framework, as displayed in Figure~\ref{fig_model}. The sample is split into its separate modalities: the textual component of the sample is fed into the BERT textual encoder, while the visual component of the sample is fed into the ResNet50 visual encoder. In the next stage of the process, BERT outputs a textual feature vector of length 768 and ResNet50 outputs a visual feature vector of length 2048. The two feature vectors are concatenated and the joint feature vector is then fed into the classification layer. The classification layer is a two-layer fully connected network which predicts the label $\hat{y}$, an estimate of the ground-truth label $y$, as formulated in Section~\ref{sec_prob}. Further architectural details are outlined below in Section~\ref{sec_setup} (under Parameter Setup).

\subsection{Experimental Setup}

This subsection discusses the experimental setup used for the investigation. This includes describing the datasets, baseline models, environment, parameters and evaluation metrics used.

\subsubsection{Datasets}

Two datasets are used for training and evaluation. The first dataset is the Fakeddit dataset by Nakamura, Levy and Wang~\cite{ref_nakamura} and represents the non-local dataset. This dataset is based on Reddit and was chosen because it is large, contains multi-domain misinformation and contains five different types of misinformation (a subset of the types outlined in Section~\ref{sec_types}). The different types of misinformation included in the Fakeddit dataset are manipulated content, false connection, satire/parody, misleading content and imposter content. For the purposes of this paper, the five types of misinformation present in the Fakeddit dataset are considered as a single misinformation class. The Fakeddit dataset used contains 3523 true samples and 2474 misinformation samples. The second dataset is a new South African specific multimodal misinformation dataset sourced from Real411\footnote{\url{https://www.real411.org/}}~\cite{ref_real411} and represents the local dataset. This dataset contains samples from various social media networks, namely: WhatsApp, Instagram, Facebook and Twitter. The Real411 dataset also contains different categories, however these categories do not align with those used in the Fakeddit dataset. The categories present in the Real411 dataset are misinformation, hate speech, incitement to violence and harassment. Since hate speech, incitement to violence and harassment do not fall under the label of misinformation, samples from these classes are excluded. The Real411 dataset used contains 378 true samples and 341 misinformation samples.

\begin{table}
\centering
\caption{Combinations of samples, from different datasets, used for training in Task 2.}\label{tab_comb}
\begin{tabular}{|P{2.5cm}|P{4cm}|P{4cm}|P{1.2cm}|}
\hline
{\bfseries Combination} &  {\bfseries Number of Samples from Fakeddit Dataset} & {\bfseries Number of Samples from Real411 Dataset} & {\bfseries Total}\\
\hline
Non-Local & 5997 & 0 & 5997 \\ \hline
Mixed & 5997 & 646 & 6643 \\ \hline
Local & 0 & 646 & 646 \\ 
\hline
\end{tabular}
\end{table}

For the purpose of Task 1 (as discussed in Section~\ref{sec_prob}), the multimodal Fakeddit dataset is used for both training and evaluation. For Task 2, three combinations of the Fakeddit and Real411 datasets are used for training, while a separate subset of the Real411 dataset is used for evaluation. The training combinations are outlined in Table~\ref{tab_comb}.

\subsubsection{Baseline Models}

The proposed MMiC model is tested on a benchmark dataset, Fakeddit, and compared with two base modalities, namely: unimodal textual and visual models. The baseline unimodal textual models include two classical machine learning models: Gaussian Naive Bayes (NB) and Logistic Regression (LR). Although considered a popular algorithm for textual classification tasks, the Support Vector Machine (SVM) model is not chosen as a baseline because results from existing misinformation detection research suggest that the NB and LR models outperform the SVM model~\cite{ref_alonso,ref_palani,ref_singhal}. SVM is also computationally expensive to implement in comparison to NB and LR, which require less computation. In addition, SVM requires a careful selection of hyperparameters which can be time-consuming. Also included as a baseline unimodal textual model is a simple BERT classifier, consisting of the textual encoder, BERT, and a classification layer. Likewise, the baseline unimodal visual model is a simple ResNet classifier consisting of the visual encoder, ResNet50, and a classification layer. 

\subsubsection{Parameter Setup}\label{sec_setup}

The experiments are carried out on a server with the specifications described in Table~\ref{tab_spec}. This setup is used for building, training and evaluating the models. Python is used to implement all code and a virtual environment isolates all the packages necessary for running the project. All models are implemented using PyTorch and Scikit-Learn libraries. The datasets are split into 80\% training and 20\% test samples. Textual pre-processing includes: converting all characters to lowercase, removing new line characters, leading and trailing spaces, URLs and punctuation. Once the text is cleaned, the textual samples are vectorized using TD-IDF (for the NB and LR models) and BERT (for the BERT and MMiC models). The pre-trained BERT model that is used is the `bert-base-uncased' model. The maximum length of the input text is set to 300 words, in order to accommodate the length of the largest textual sample. The pre-trained ResNet model that is used us the `microsoft/resnet-50' model. The input images are normalized and resized to a dimension of 560 x 560. All models are trained for a maximum of 10 epochs, however an Early Stopping optimization technique is used to select the actual number of epochs in order to prevent over-fitting. Thus, the model from the epoch which achieves the best validation accuracy is retained. The cross-entropy loss function and AdamW optimizer are used to train the models. The learning rates for the BERT, ResNet and MMiC classifiers are set to $1 \times 10^{-5}$, $1 \times 10^{-3}$ and $5 \times 10^{-6}$, respectively.

\begin{table}
\centering
\caption{Server Specifications}\label{tab_spec}
\begin{tabular}{|P{2.5cm}|P{6cm}|}
\hline
{\bfseries Component} &  {\bfseries Specification}\\
\hline
OS &  Ubuntu Server 20.04   \\ \hline
CPU & 12th Gen Intel(R) Core(TM) i9-12900 \\ \hline
RAM (CPU) & 125GB DDR4 \\ \hline
GPU & 2 x NVIDIA RTX A6000 \\ \hline
RAM (GPU) & 2 x 50GB \\ \hline
Disk Size & 2TB after RAID \\
\hline
\end{tabular}
\end{table}

\subsubsection{Evaluation Metrics}

Traditional machine learning performance metrics are used for evaluation. These metrics originate from the confusion matrix and include the terms (TP, TN, FP and FN) which stand for true positive, true negative, false positive and false negative, respectively. The first metric used is a measurement of the model's overall accuracy when classifying misinformation, as shown in Equation~\ref{equ_2}:

\begin{equation}
Accuracy = \frac{TP + TN}{TP + TF + FP + FN}
\label{equ_2}
\end{equation}

The second metric used is precision, which is a measurement of the model's ability to correctly predict misinformation, as shown in Equation \ref{equ_3}:

\begin{equation}
Precision = \frac{TP}{TP + FP}
\label{equ_3}
\end{equation}

The third metric used is recall, which  is a measurement of the model's ability to identify all cases of misinformation, as shown in Equation~\ref{equ_4}:

\begin{equation}
Recall = \frac{TP}{TP + FN}
\label{equ_4}
\end{equation}

The final metric used is the F1-score, which takes into account both the model's precision and recall, and is shown in Equation~\ref{equ_5}:

\begin{equation}
F1\text{-}Score = \frac{2 \times Precision \times Recall}{Precision + Recall}
\label{equ_5}
\end{equation}

\section{Results}

The results of the proposed MMiC model, as well as the results of the unimodal baseline models, for Task 1 and Task 2 are displayed in Table~\ref{tab_res1} and Table~\ref{tab_res2}, respectively. For each task, the experiment is repeated five times and the mean result recorded. Furthermore, there are two phases involved during the evaluation on each task, namely: cross validation (CV) and test. The Hold Out CV approach is used, as this is computationally inexpensive when compared to other CV techniques. During CV, the models are trained and tested on random portions of the datasets (i.e., for each repetition of the experiment, the train and test sets are different). During the test phase, the models are trained and tested on the same portions of the datasets (i.e., for each repetition of the experiment, the train and test sets remain unchanged). The CV phase is used to validate the results collected during the test phase.

\begin{table}
\centering
\caption{F1-Scores of models trained and evaluated on Fakeddit dataset. Standard deviations are shown in parenthesis, for CV. }\label{tab_res1}
\begin{tabular}{|P{0.5cm}|P{2.3cm}|P{2.3cm}|P{2.3cm}|P{2.3cm}|} \hline
 & {\bfseries Model} &  {\bfseries Binary} &  {\bfseries Multi-3} &  {\bfseries Multi-6} \\ \hline
\multirow[c]{5}{*}[0mm]{\rotatebox{90}{\bfseries CV}}
 & NB & 0.68 (0.02) & 0.58 (0.01) & 0.45 (0.01) \\ 
 & LR & 0.74 (0.01) & 0.75 (0.02) & 0.62 (0.01) \\ 
 & BERT & 0.83 (0.01) & 0.81 (0.02) & 0.70 (0.02) \\ 
 & ResNet & 0.74 (0.01) & 0.73 (0.02) & 0.70 (0.03) \\ 
 & MMiC & {\bfseries 0.85} (0.01) & {\bfseries 0.85} (0.02) & {\bfseries 0.83} (0.01) \\ \hline\hline
\multirow[c]{5}{*}[0mm]{\rotatebox{90}{\bfseries Test}}
 & NB & 0.69 & 0.57 & 0.43 \\ 
 & LR & 0.76 & 0.75 & 0.62 \\ 
 & BERT & 0.86 & 0.85 & 0.72 \\ 
 & ResNet & 0.75 & 0.73 & 0.71 \\ 
& MMiC & {\bfseries 0.88} & {\bfseries 0.88} & {\bfseries 0.83} \\ \hline
\end{tabular}
\end{table}

\begin{figure}[htb]
  \centering
  \includegraphics[width=\textwidth]{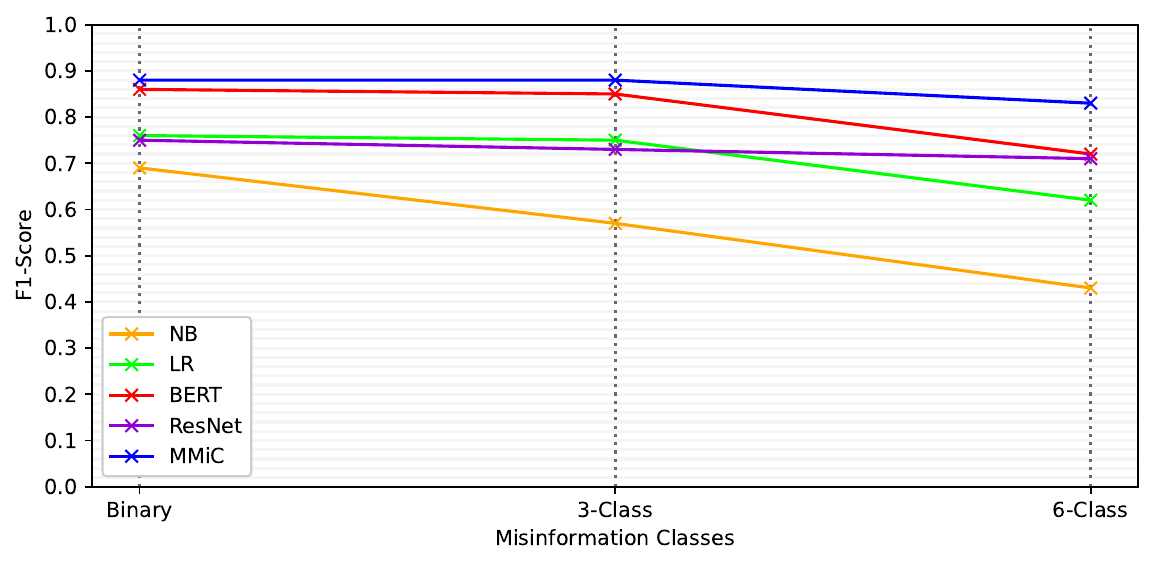}
  \caption{Line graph of results from Task 1.} 
  \label{fig_res1}
\end{figure}

The results in Table~\ref{tab_res1} highlight that the MMiC model matches the performance of existing state-of-the-art misinformation detection models on the benchmark dataset, Fakeddit. The MMiC model also outperforms all other tested models across the different classification settings for both the CV and test phases. MMiC achieves top performance during the test phase: 88\% for binary, 88\% for 3-class and 83\% for 6-class classification settings. The BERT model is second to the MMiC model, followed by the ResNet model. The LR model shows similar performance to the ResNet model for binary (76\%) and 3-class (75\%) classification settings but drops off significantly on the 6-class classification setting (62\%). The NB model performs the worst out of all the models in this task: 69\% for binary, 57\% for 3-class and 43\% for 6-class classification. The fact that BERT outperforms the ResNet classifier may suggest that the textual component of the misinformation samples provide more information (and therefore, more evidence) to the classification process, than the visual components of the samples. In Figure~\ref{fig_res1}, it can be seen that as the number of classification classes increases, the performance of the models decrease. However, as the number of classification classes increases, the MMiC model performs significantly better than the other models. This may suggest that multimodal models are less vulnerable to performance decay as the classes in the misinformation classification task increases.

The results in Table~\ref{tab_res2} illustrate that the MMiC model performs the best when trained on local samples to classify local misinformation (88\%). It is also observed that the MMiC and BERT show similar performance, outperforming all other models across all different training environments, during the CV and test phases. The fact that the MMiC and BERT models show similar performance on this task may be a result of the images in the Real411 dataset not adding significant information to the classification process. This may be a consequence of many Real411 images being screenshots of social media posts (i.e., textual content).

\begin{table}
\centering
\caption{F1-Score of models trained on combination of local and non-local samples. Standard deviations are shown in parenthesis, for CV.}\label{tab_res2}
\begin{tabular}{|P{0.5cm}|P{2.3cm}|P{2.3cm}|P{2.3cm}|P{2.3cm}|} \hline
 & {\bfseries Model} & {\bfseries Non-Local} &  {\bfseries Mixed} & {\bfseries Local}\\ \hline
\multirow[c]{5}{*}[0mm]{\rotatebox{90}{\bfseries CV}}
 & NB & 0.53 (0.02) & 0.55 (0.02) & 0.54 (0.03) \\ 
 & LR & 0.50 (0.04) & 0.59 (0.04) & 0.63 (0.05) \\ 
 & BERT & {\bfseries 0.60} (0.08) & 0.83 (0.02) & 0.84 (0.02) \\ 
 & ResNet & 0.54 (0.05) & 0.57 (0.02) & 0.57 (0.02) \\ 
 & MMiC & 0.54 (0.03) & {\bfseries 0.84} (0.02) & {\bfseries 0.85} (0.03) \\ \hline\hline
\multirow[c]{5}{*}[0mm]{\rotatebox{90}{\bfseries Test}}
 & NB & 0.55 & 0.60 & 0.59 \\ 
 & LR & 0.57 & 0.59 & 0.64 \\ 
 & BERT & {\bfseries 0.63} & {\bfseries 0.87} & 0.87 \\ 
 & ResNet & 0.54 & 0.59 & 0.61 \\ 
& MMiC & 0.53 & 0.86 & {\bfseries 0.88}\\ \hline
\end{tabular}
\end{table}

\begin{figure}[htb]
  \centering
  \includegraphics[width=\textwidth]{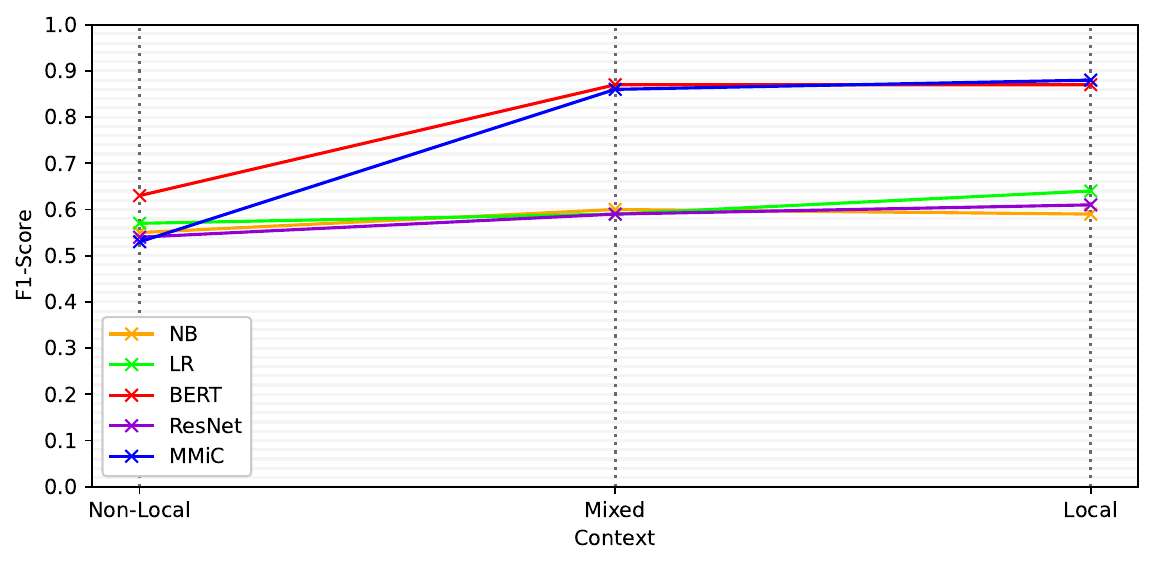}
  \caption{Line graph of results from Task 2.} 
  \label{fig_res2}
\end{figure}

From the results in Table~\ref{tab_res2}, it is observed that all models appear to increase in performance when local misinformation samples are introduced into the training dataset. This trend can also be observed in Figure~\ref{fig_res2} and highlights the importance of incorporating local samples into the training process of a MD model. Furthermore, most models (except NB) perform the best when trained in a local contextual environment. However, the mean performance increase of the models between a mixed and local training environment is 1\%. Whereas, the mean performance increase of the models between a non-local and mixed training environment is 14\%. The total mean performance increase of the models between a non-local and mixed or local training environment is 15\%. Therefore, the results show that it is important to include local misinformation samples when training a misinformation detection model to detect misinformation in a local environment. The effect of including local samples is especially influential for the BERT and MMiC models (as observed in Figure~\ref{fig_res2}) resulting in a mean performance increase of 29\% between a non-local and mixed or local training environment. Overall, it should be noted that the training dataset does not necessarily need to only include local samples in order to significantly improve performance. The results observed provides evidence supporting the potential influence of concept drift, as discussed in Section~\ref{sec_concept}.

\section{Conclusion}

In this article, a multimodal misinformation detection model (called MMiC) is developed and a new misinformation dataset (called Real411), which is set in the South African social media environment, is introduced. MMiC makes use of both textual and visual information to classify misinformation by means of a BERT and ResNet encoder. The results show that MMiC matches the performance of current state-of-the-art misinformation detection models on the Fakeddit dataset. Furthermore, multimodal models decay to a lesser degree than unimodal models when the number of misinformation classes are increased. The results also show that including local misinformation samples into the training of a misinformation detection model significantly improves the model's ability to detect misinformation in a local contextual environment. Future studies include investigating the performance of different deep learning models within the MMiC framework on the Real411 dataset. Although Real411 is a predominately English dataset, South Africa has 11 national languages, which may influence misinformation detection in the South African social media environment.

\subsubsection{Acknowledgements} 

We want to acknowledge Real411, a program of Media Monitoring Africa, for the dataset. We would like to acknowledge funding from the ABSA Chair of Data Science, Google and the NVIDIA Corporation hardware Grant.

%
%
%
\bibliographystyle{splncs04} %
\bibliography{refs}
%
%
\end{document}